\documentclass[journal,transmag]{IEEEtran}
\usepackage{lineno,hyperref}
\usepackage{graphicx}
\usepackage{color}
\usepackage[section]{placeins}
\usepackage{float}
\usepackage{dblfloatfix}
\ifCLASSINFOpdf
\else
\fi
\begin{document}
%
\title{Multi-scale predictions for robust hand detection and classification}


\author{\IEEEauthorblockN{Lu Ding\IEEEauthorrefmark{1},
Yong Wang\IEEEauthorrefmark{2},
Robert Lagani$\grave{e}$re\IEEEauthorrefmark{2},
Xinbin Luo\IEEEauthorrefmark{3}, and
Shan Fu\IEEEauthorrefmark{3}}
\IEEEauthorblockA{\IEEEauthorrefmark{1}School of Aeronautics and Astronautics, Shanghai Jiao Tong University, Shanghai, China}
\IEEEauthorblockA{\IEEEauthorrefmark{2}School of Electrical Engineering and Computer Science, University of Ottawa, Ottawa Canada}
\IEEEauthorblockA{\IEEEauthorrefmark{3}School of Electronic information and Electrical Engineering, Shanghai Jiao Tong University, Shanghai, China}
\thanks{Corresponding author: Yong Wang (email: ywang6@uottawa.ca).}}

\markboth{Journal of \LaTeX\ Class Files,~Vol.~14, No.~8, August~2015}%
{Shell \MakeLowercase{\textit{et al.}}: Bare Demo of IEEEtran.cls for IEEE Transactions on Magnetics Journals}
%



\IEEEtitleabstractindextext{%
\begin{abstract}
In this paper, we present a multi-scale Fully Convolutional Networks (MSP-RFCN) to robustly detect and classify human hands under various challenging conditions. In our approach, the input image is passed through the proposed network to generate score maps, based on multi-scale predictions. The network has been specifically designed to deal with small objects. It uses an architecture based on region proposals generated at multiple scales. Our method is evaluated on challenging hand datasets, namely the Vision for Intelligent Vehicles and Applications (VIVA) Challenge and the Oxford hand dataset. It is compared against recent hand detection algorithms. The experimental results demonstrate that our proposed method achieves state-of-the-art detection for hands of various sizes.
\end{abstract}

}

\maketitle

\IEEEdisplaynontitleabstractindextext

%
\IEEEpeerreviewmaketitle

\section{Introduction}
Robust hand detection is a fundamental task in human-computer interaction and in human activity monitoring. Many studies have been devoted to the hand detection and classification problem [1, 2, 3, 4]. However, in spite of these recent progresses, there are still a large number of challenges that need to be overcome in practical deployments. Hands in an image are generally small, low resolution objects. They are subject to occlusion and illuminations changes and more importantly, hands are inherently deformable objects that appear in a large diversity of poses. Figure 1 illustrates some hand detection results in challenging situations as obtained by our proposed approach.

Traditional methods [5, 6] use hand-craft features to encode the human hand shape and appearance. The limited discriminative capabilities of these approaches make difficult the reliable detection of human hands in complex situations. More recently, convolutional neural networks (CNNs) have achieved appealing results in image classification [7, 8], object detection [9, 10, 11], image segmentation [12], pedestrian detection [13] and so on.

In particular, the Faster R-CNN framework and its extensions achieve the state of the art results on several benchmark datasets.
A fully convolutional network is employed in RPN to generate a set of object proposals which include object boundaries and objectness scores. Anchors with different sizes and shapes are employed to guarantee translation invariance. The R-CNN method then uses a CNN to classify these proposals.

Despite their success, these approaches have certain drawbacks that make them less flexible. Firstly, Faster R-CNN does not very well handle small objects. This is mainly because only the last convolutional layer of a deep convolutional network is used to extract region proposals. Taking VGG16 as an example, the last convolutional layer is $Conv5\_3$ which is 16 times smaller than the input image. This means that, given a hand region smaller than 16*16 pixels, the projected region of interest (RoI) for that location would be less than one cell in size. As a result, small region proposals can not be produced. In addition, the misalignment in the RoI pooling severely affects small object localization [26].

Secondly, CNNs for detection are very often constructed from model built for image classification. These fully convolutional networks are designed to favor translation invariance for image classification. This translational invariance can significantly impact the detection accuracy of object detections built on top of these networks.
\begin{figure*}
\begin{center}
\includegraphics[width=1\textwidth]{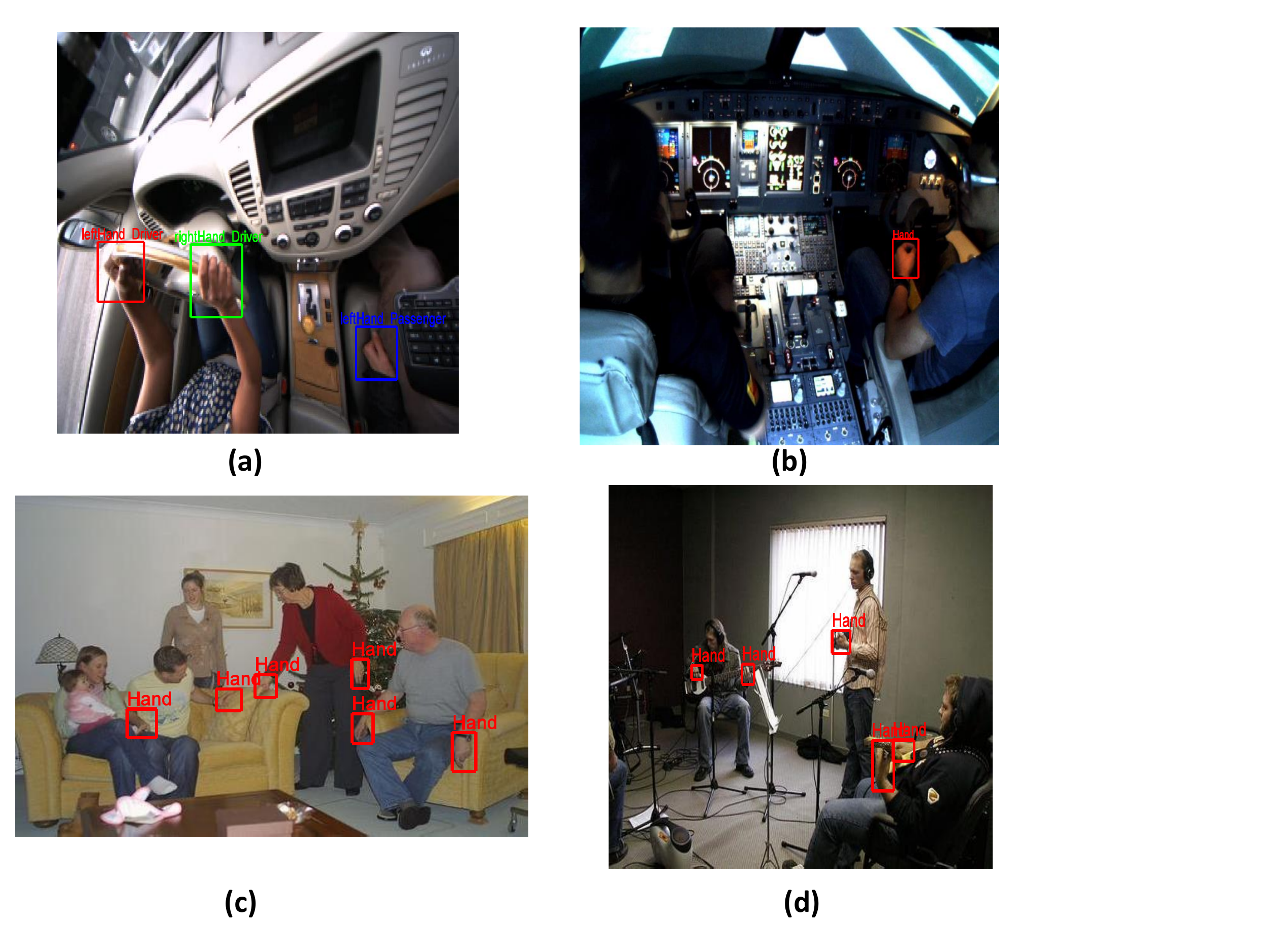}
\caption{Some examples of hand detection results using our proposed method. Our proposed method can effectively and robustly detect hands across occlusions, illumination variations and low resolution conditions. (a) Different colors represent different hand classification.}
\label{fig:1}       
\end{center}
\end{figure*}
The goal of this paper is to leverage the pyramidal architecture of CNNs to generate dense region proposals from which multi-scale predictions for hand detection and classification are produced. To achieve our goal, we propose a simple but effective method using feature maps of different scales to generate class prediction and localization of hands. Rather than building up feature pyramid by concatenating or adding feature maps from different scales together, we use each layer of feature maps in the CNNs to generate class scores and to regress bounding boxes positions. The prediction made by higher level feature maps contains stronger contextual semantics while the lower level ones integrate more localized information at finer spatial resolution. These predictions are fused together to make final decisions. We use this multi-scale prediction network for RPN and region fully convolutional networks (RFCN) [11] to perform hand detection.

In summary, our contributions are as follows,

(1) A multi-scale prediction scheme is employed in the RPN and RFCN stages to generate multi-scale prediction for hand detection and classification.

(2) The proposed method is tested on the ARJ dataset in which hand images have been collected in difficult conditions and achieved appealing performance. In addition, experiments are presented on two challenging hand datasets, the Vision for Intelligent Vehicles and Applications (VIVA) Challenge [14] and Oxford Hand Detection database [15].

The rest of this paper is organized as follows. In section 2, we review prior work on object detection, and analyze their limitations in the context of hand detection and classification. Our proposed approach for detecting and classifying hands is given in section 3. Section 4 presents experimental results on hand datasets. Finally, our conclusions on this work are drawn in section 5.
\section{Related work}
In this section, we first review prior methods in object detection. Then, we present works related to the hand detection problem. Finally, some limitations of the R-FCN framework when applied to the hand detection problem are discussed.

One of the most important approaches in object detection is the family of Region-based CNNs. The first generation of this family is R-CNN [16] which applies a CNN to classify given region proposals. The CNN is employed as a feature extractor and is further trained for object detection with Support Vector Machines (SVM). Then, bounding-box regression is applied. This approach achieves very high accuracy but is also very time consuming.

Fast R-CNN [9] shares features between proposals to reduce the computation time. The network has been developed such to compute a feature map only once per image in a fully convolutional fashion. RoI pooling is then employed to dynamically sample features from the feature map for each object proposal. A multi-task loss is also used in the network, i.e. a classification loss and a bounding-box regression loss. Even if the RoI pooling layer has the benefit of accelerating the detection, the region proposal step, designed outside of the network, still remains a bottleneck. It results sub-optimal solutions and high dependence on the external region proposal methods.

With Faster R-CNN [10] a region proposal network is implemented in a fully convolutional manner to estimate object bounding boxes and scores. Proposal anchors are included at different scales and ratios to achieve translation invariance. The RPN shares its convolution features with the detection network. Therefore proposal generation and object detection are computed very efficiently using very deep VGG-16 model [7].

R-FCN [11] consists of two stages corresponding to region proposal and region classification. R-FCN includes shared, fully convolutional architectures similar to its predecessor FCN [12]. The last few fully connected layers are replaced by convolutional layers to make end-to-end learning and inference more efficient. To solve the dilemma between translation invariance in image classification and translation-variance in object detection, R-FCN employs a group of convolutional layers at the FCN output to generate position sensitive score maps. In their work, Residual Networks [8] are adopted for object detection. The average pooling layer and the fully connected layer is removed and only the convolutional layers are used to compute feature maps.

The Region-based CNN methods, e.g. R-CNN, Fast R-CNN, Faster R-CNN [16, 9, 10] and the recent R-FCN [11] achieve state-of-the-art performance in object detection. These methods can detect objects with very high accuracy. However, these methods are less effective when it comes to detect very small objects. In hand detection dataset, hand regions are often small (Figure 1).
In addition, both RPN and R-FCN make predictions based on one single high-level convolutional feature map. For example, in ResNets-101, $Conv4\_{23}$ layer is a high-level feature with a spatial resolution that is 16 times smaller compared to the input image. Consequently, only several pixels information can be send to $Conv5\_x$ layer. Therefore, the detector will have much difficulty to predict the object class and bounding box location based on information from only few pixels.
\section{Our Approach}
\subsection{Overview of Our Approach}
This section presents our proposed multi-scale prediction RFCN (MSP-RFCN) method to effectively and robustly detect and classify hand regions. Our method utilizes CNNs features encoded in both the global and the local representation for hand regions. Figure 2 illustrates the pipeline of our hand detection and classification framework.

Our dense RPN extract, from the input image, proposals of different scales. Next, features in different layers are employed for the multi-scale prediction. Finally, hands of different sizes are detected and classified.

\begin{figure*}
\begin{center}
  \includegraphics[width=1\textwidth]{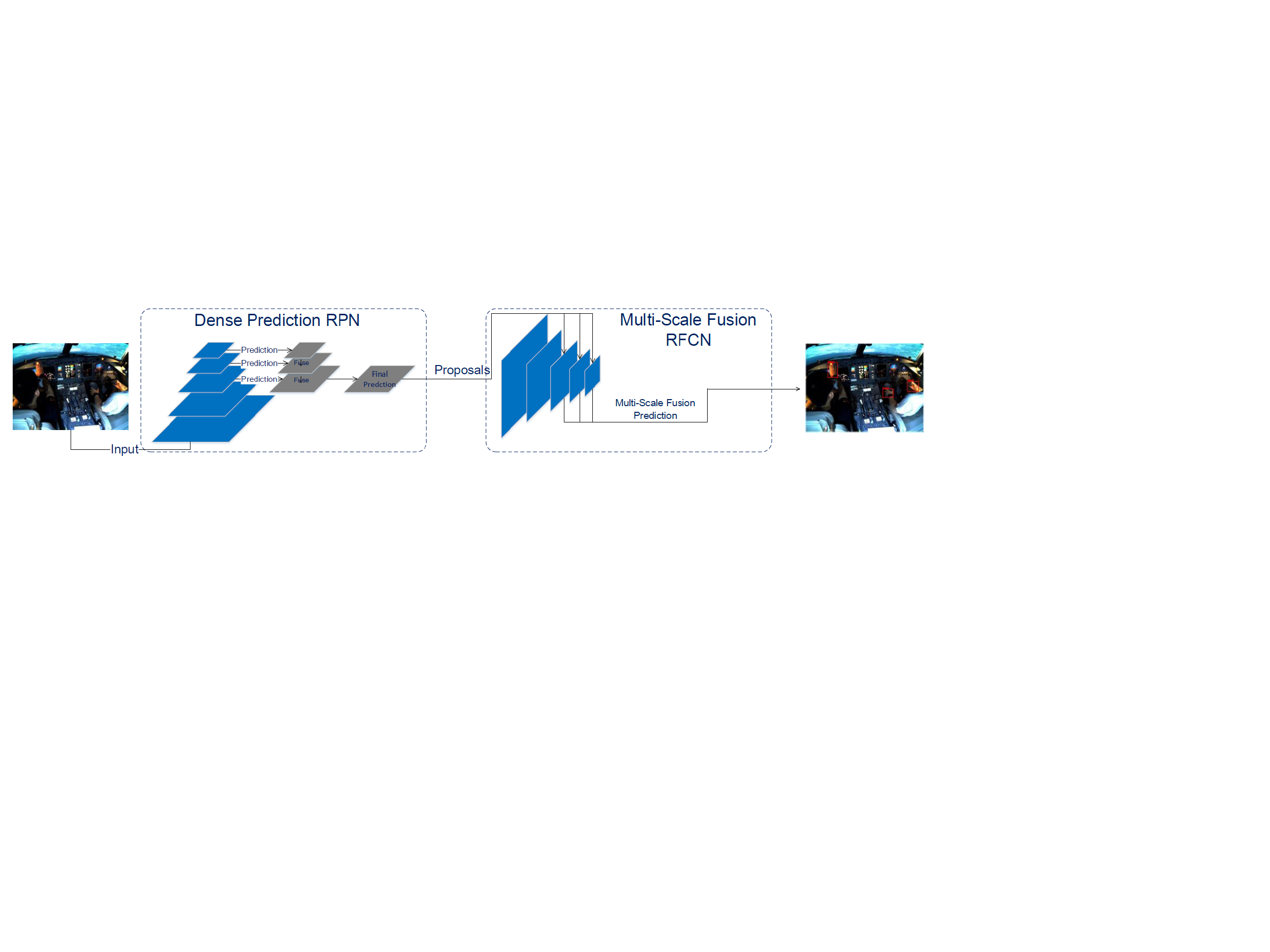}
\caption{The illustration of our hand detection and classification framework.}
\label{fig:2}       
\end{center}
\end{figure*}
\subsection{Deep Network Architecture}
Human hands in images are usually captured under challenging situations, such as low resolution, viewpoint changes, illumination variation, to name a few. It is not easy for an R-FCN to detect these hand regions since the receptive field in the last convolution layer is quite large. Since in a hierarchical network, there is strong semantic meaning in high-level feature maps and finer-resolution information in low-level feature maps, our idea is to fuse feature map of different scales to make final decisions. Therefore, a combination of multiple scales with global and local features is able to enhance performance in hand detection and classification.
Specifically, the pre-trained ResNets-101 model [8] is adopted in our work. There are 5 major parts, i.e. $Conv1$, $Conv2$, $Conv3$, $Conv4$ and $Conv5$, followed with pooling layers which shrink the spatial scale. For our model, we consider the convolution layers $\{Conv3\_4, Conv4\_{23}, Conv5\_3\}$, applying ��hole algorithm�� [12] to ensure they have same spatial resolution. We apply strides of $\{4, 8, 16\}$. These convolution layers are then normalized (by $L_2$ norm) and concatenated together as the feature map for region proposal.

Figure 3 shows the architecture of our network. C1, C2, C3, C4 and C5 represent the final output of Conv1, Conv2, Conv3, Conv4 and Conv5 respectively. We do not include $Conv1$ and $Conv2$ into the pyramid due to its large memory footprint and 4 strides is enough for object detection. More importantly, $Conv1$ and $Conv2$ contain too much low level shape details and they are still too shallow to abstract information for prediction. A prediction head which is a 3*3 convolutional layer followed by two sibling 1*1 convolutions is attached to the feature maps output by $Conv5$, $Conv4$ and $Conv3$. We adopt a two-stage object detection strategy which consists of i) region proposal and ii) region classification. Our proposed multi-scale prediction scheme is extended from RPN and R-FCN to generate dense region proposals and detect small hands. For RPN, we have to deal with different resolution of the feature maps when we fuse different scale predictions. And for R-FCN, score map of each region proposal output in each stage is of the same size via position sensitive RoI pooling (PSROIPooling). Next, we will describe our region proposal and region classification in details.
\begin{figure*}
\begin{center}
  \includegraphics[width=1\textwidth]{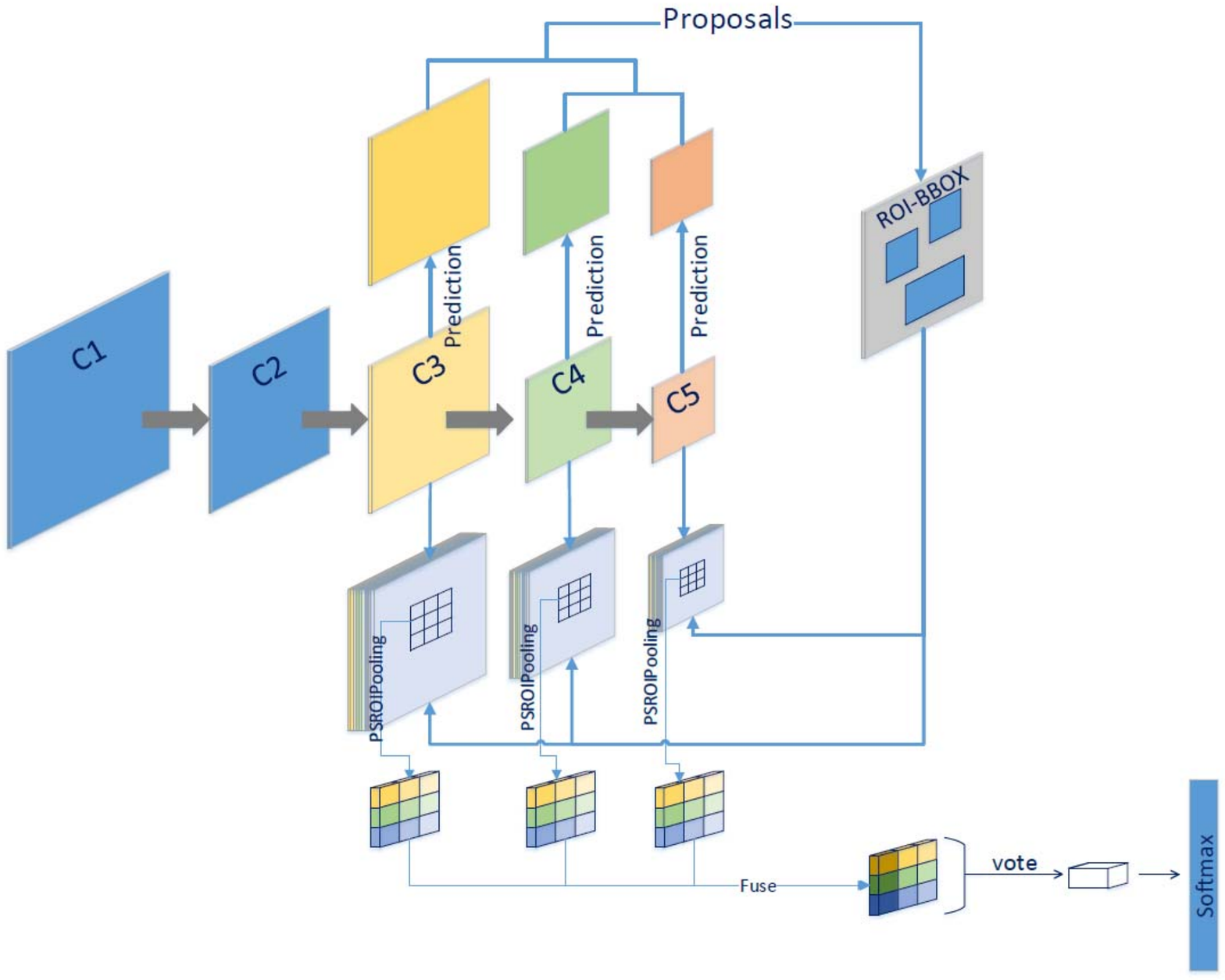}
\caption{Architecture of our multi-scale predictions RFCN. C1, C2, C3, C4 and C5 represent the final output of Conv1, Conv2, Conv3, Conv4 and Conv5 respectively.}
\label{fig:3}       
\end{center}
\end{figure*}
\subsection{Region Proposal Network}
RPN is a fully convolutional network that uses a sliding window and class-agnostic detection to classify foreground object from background. In the original RPN design, a small sub-network is evaluated on dense 3*3 sliding windows on the $C\_5$ with a feature stride of 16. The small sub-network performs object/non-object binary classification and bounding box regression on top of a single scale convolutional feature map.

The object/non-object criterion and bounding box regression target are defined with respect to a set of reference boxes called anchors. The anchors are of multiple pre-defined size and aspect ratios in order to cover objects of different shapes. For small objects, anchors located at layers where strides are 16 are still too sparse. Since layer of lower levels sees fewer pixels, it makes sense that finer predictions should be made on the finer resolution but shallower layer.

The anchor distribution can be made denser by combining predictions from different layers. This is done by introducing skip connections [27] that are used to fuse predictions by element-wise addition. The predictions come not only from the previous branch but also from branches directly connecting previous lower-level layers.
\begin{figure*}
\begin{center}
  \includegraphics[width=1\textwidth]{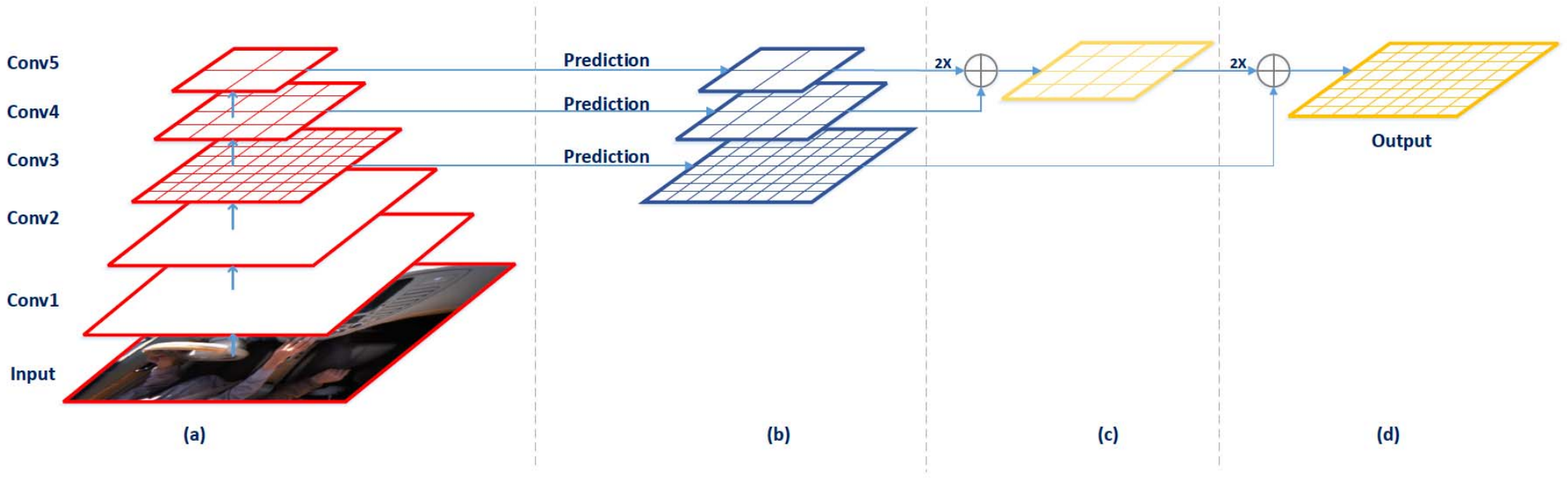}
\caption{Our proposed dense region proposal network. (a) A feed forward propagation is carried out to an input image to generate hierarchical features; (b) A prediction head is attached to $C\_3$, $C\_4$ and $C\_5$; (c) and (d) up-sampling the coarse prediction and fuse with finer predictions by element-wise addition. The final output is 4 times finer than origin region prediction, which merges high-level prediction and low-level prediction.}
\label{fig:3}       
\end{center}
\end{figure*}

A prediction head is attached to $C\_3$, $C\_4$ and $C\_5$. Then 3 scales of prediction are made at the scaling step of 2, which is denoted as $P\_3, P\_4, P\_5$.
Next $P\_3$ is up-sampled by a factor of 2. And the up-sampled prediction is merged with $P\_4$ by element-wise addition to get a prediction which is twice finer than the original one, denoted as $P\_{34}$. Then we do the same trick to fuse $P\_{34}$ with $P\_5$ to obtain the final prediction (see Figure 4).
There are many ways to perform up-sampling such as $\grave{a}$ trous algorithm [12], nearest neighbor and interpolation. In our work, we adopt the backwards stride convolution which is a more efficient and effective up-sampling approach [25]. Up-sampling by a factor $f$ corresponds to a convolution with a fractional input stride of $\frac{1}{f}$. A natural way to up-sample is to perform de-convolution with an output stride of f. Using de-convolution is efficient and easy to implement since it simply reverses the forward and backward passes of convolution. Unlike $\grave{a}$ trous algorithm and shift-and-stitch method [12], up sampling is performed within the network, thus we can train the network end-to-end. We initialize the de-convolution weight filer to perform bilinear interpolation and unfreeze the filter to learn by back propagation from the pixel wise loss function.
\FloatBarrier
\subsection{Region Classification}
\label{sec:intro}
In our region classification procedure, the network model is similar to R-FCN which ends with a position-sensitive RoI pooling layer. This layer aggregates the outputs of the last convolutional layer and produces scores for every RoIs which are classified into object categories or background. In our model, all weighted layers are convolutional and are computed on the entire image. Through end-to-end training, specialized position-sensitive score maps are learned by the RoI layers. At the last convolutional layer, a group of $k*k$ position-sensitive score maps of each class is generated, and thus a $k^2(C+1)$- channel output layer with C classes is produced. Selective pooling is carried out by the position-sensitive RoI layer. Each of the $k*k$ bin aggregates responses from only one score map out of the group of $k*k$ score maps. Since the receptive field of $C\_5$ is quite large and the RoI is small, the features in $Conv5$ contain high-level information. Hence, the score maps may not be activated at the relative position of an object, which will lead to false negative. Since the lower layer (e.g., $Conv3$, $Conv4$) see fewer pixels, the position sensitive score map for small objects should be computed at the shallower layer with finer resolution feature maps. By merging the position sensitive score maps of shallow and deep layers together, false positive detection can be reduced. The bounding box regression is then performed following the same process as classification.

Figure 5 shows our proposed network. At each level, we compute a $k^2*(C + 1)$-channel position sensitive score maps. In this work we use $C\_3$, $C\_4$ and $C\_5$ output by $Conv3$, $Conv4$ and $Conv5$ with the stride of {4, 8, 16} (as we modified the origin ResNets to enlarge feature map resolution). We perform position sensitive RoI pooling at each score maps computed at $Conv3$, $Conv4$ and $Conv5$. Though the receptive fields are different in these layers, the output of position sensitive RoI pooling are all of the same size. Next the predictions of each layer are merged by element-wise addition. And consequently small size objects can be detected.

\begin{figure*}
\begin{center}
  \includegraphics[width=1\textwidth]{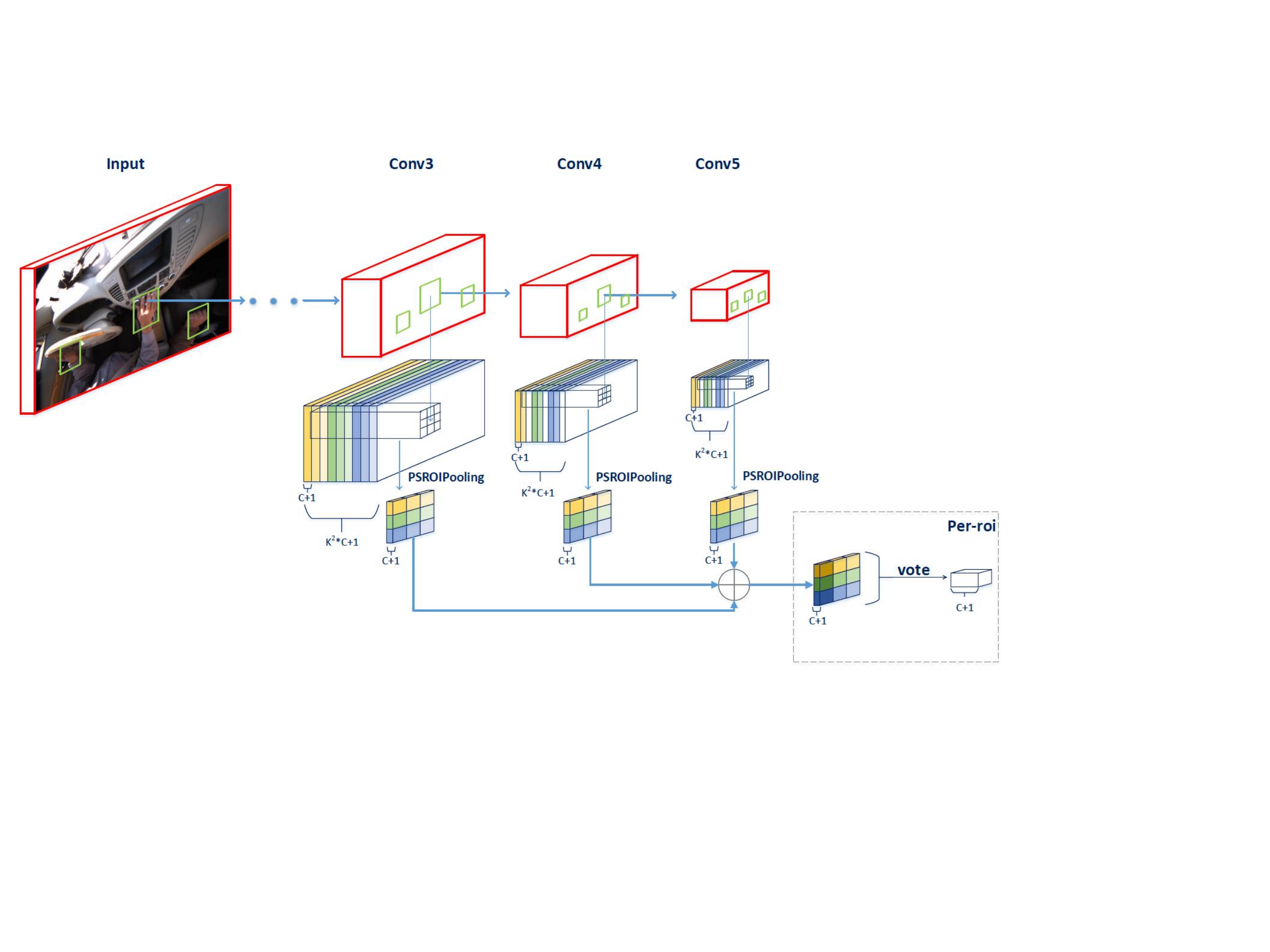}
\caption{Illustration of region classification procedure. Networks with skip connection combine predictions made by coarse feature maps and finer feature maps. Predictions are fused at the output, leveraging both detail and global information.}
\label{fig:4}       
\end{center}
\end{figure*}
\subsection{Implementation}
Feature maps output from different layers greatly vary in amplitude. Therefore, prediction from different layers cannot be simply concatenated [21]. Based on our observation, the deeper a layer is, the smaller the values of feature map pixels are. This phenomenon leads to prediction made by lower level feature map to dominate the final decision. In order to address this issue, a $L_2$ normalization layer [21] is added after $C\_3$, $C\_4$ and $C\_5$ respectively. The normalization and scaling operation is performed element-wise along each channel axis.

Here, we briefly introduce our multiple feature map prediction process which is based on the R-FCN. More details can be found in [11].
Each RoI is divided into $k\times k$ bins by a regular grid to encode position information. For an RoI rectangle of a size $w \times h$, a bin is of the size $ \approx \frac{w}{k}\times \frac{h}{k}$. Inside the $\left(i,j\right)$-th bin $( 0 \leq i,j \leq k)$, a position-sensitive RoI pooling operation is employed to pool over the  $\left(i,j\right)$-th score map:
\begin{eqnarray}
r_{c}\left ( i,j\mid \Theta   \right ) = \sum_{l\in{P_3,P_4,P_5}}\sum_{\left(x,y \right )\in bin\left ( i,j \right )}\nonumber\\
z_{i,j,c,l}\left ( x+x_{0},y+y_{0} \right) \mid \Theta )/n
\end{eqnarray}
where $r_{c}\left ( i,j\mid \Theta   \right )$ is the layer aggregated pooled response in the $\left( i,j \right)$-th bin for the c-th category, $z_{i,j,c,l}$ is one score map in the $k^2(C+1)$ score maps outputted by one layer feature map. $\left(x_0,y_0 \right)$ is the top-left corner of an RoI, $n$ is the number of pixels in the bin, and $\Theta$ is all the learnable parameters of the network. The response of the RoI score is pooled inter-layer and fuse together across the layer by element-wise addition.

Then $k^2$ position-sensitive score votes on the RoI are performed. We aggregate the votes on an RoI to generate a $(C+1)$-dimensional vector and feed to a softmax classification across the categories. Next, a cross entropy loss is used to evaluate loss during online training and to rank the RoI during inference.

Bounding box regression is done in a similar way. We attach a sibling $4k^2$-d convolutional layer for bounding box regression after each feature map output by $\{Conv3,Conv4,Conv5\}$ aside from the $k^2(C+1)$-d convolutional layer for classification.

Our training and inference schemes are similar to the work [11] while feature maps of $\{Conv3,Conv4,Conv5\}$ are employed in our work.
\section{Experiment}

\subsection{Implementation Details}
\subsubsection{Dataset}
The VIVA Challenge [14] includes 2D bounding boxes around hands of drivers and passengers from 54 video sequences collected in driving settings of hand movements, illumination variation, and common occlusion. There are 5,500 training and 5,500 testing images.

The Oxford hand dataset [15] includes images collected from various public image dataset. In each image, all the hands that are clearly perceived by humans are annotated. For this dataset, hand regions larger than a fixed area (1500 sq. pixels) are used for testing. This gives 4,170 high quality hand instances. A training dataset is constructed from six sources including INRIA pedestrian, Buffy Stickman, Skin Dataset, Poselet (H3D), PASCAL 2007, PASCAL 2012 with 9,163 hand instances and 2,863 large hand instances. The testing dataset includes 1,856 hand instances and 649 large hand instances from movies. The validation dataset includes 2,031 hand instances and 660 large hand instances from PASCAL 2007 and PASCAL 2012.

The ARJ dataset has been collected by ourselves in a simulated airplane situations. The dataset includes 2D bounding boxes around hands of pilots from 14 video sequences collected in settings of hand movements, motion blur and occlusion. There are 287 testing images. The situation is in low lighting condition in order to not affect the operation of the pilots.
\subsubsection{Evaluation Methods}
The average precision (AP), average recall (AR) rate, and frame per section (PFS) is employed to evaluate the performance on the VIVA dataset. AP is defined as the area under the Precision-Recall curve. AR is computed over 9 evenly sampled points in log space between $10^{-2}$ and $10^0$ false positives per image.
Detection result is considered true or false according to its overlap with the ground-truth bounding box. A box is positive if the overlap score is more than a threshold (0.5). The overlap score between two boxes is defined as $\frac{GT\cap DET}{GT\cup DET}$, where GT is the axis aligned bounding rectangle around area ground truth bounding box and DET is the axis aligned rectangle around detected bounding box. The VIVA challenge is evaluated on two levels: Level-1 (L1): hand instances with minimum height of 70 pixels, only over the shoulder (back) camera view. Level-2 (L2): hand instances with minimum height of 25 pixels, all camera views. For both of the evaluation metrics, higher is better.
Classification is composed of left-right (L-R) hand classification, driver-passenger (D-P) hand classification.

The deep learning Caffe framework [24] is utilized in our implementation. Two CNNs network ResNets-101 [8] and VGG16 [7] are employed. The proposed method is evaluated on a 64 bits Ubuntu 14.04 computer with a single NVIDIA GeForce GTX TITAN X GPU and 12GB memory.
\subsection{Experiments on VIVA Dataset}
Figure 6 demonstrates the ROC and FPPI of our method and the most recent methods FRCNN+VGG [22], MS-FRCNN [3], FRCNN+Context [22], FRCNN [17], ACF Depth4 [18], YOLO [23], CNNRegionSampling [1], MS-RFCN [4]. Ours (VGG16) and Ours (ResNets-101) are our MSP-RFCN method with two different network models.

Table 1 summaries the performance of AR and AP and FPS at both levels (L1 and L2). Compare to the second best methods MS-RFCN, our proposed approach (Ours (ResNets-101)) improves the L1-AP by 0.6\% whereas the AR obtained is imprhoved by 9.3\% and 4.2\% on L1-AR and L2-AR, respectively. The processing time of our proposed method is at 5.0 FPS on GPU.

Table 2 summaries the performance of left-right hand and driver-passenger hand of the most recent hand classification approaches FRCNN+VGG [22], ACF Depth4 [18], CNNRegionSampling [1], MS-RFCN [4] and our proposed MSP-RFCN method. Compare to MS-RFCN, our proposed method (Ours (ResNets-101)) achieves state-of-the-art results with improvements of 6.0\% for AP and 3.1\% for AR on the left/right hand classification and 10.8\% for AP and 7.8\% for AR on the driver/passenger hand classification on the VIVA database. From Figure 7 and Figure 8, we can see that our proposed MSP-RFCN outperforms other methods on AP, AR and processing time.

Figure 7 demonstrates some examples of hand detection and classification by our proposed MSP-RFCN method on VIVA dataset.

\begin{figure*}
\begin{center}
  \includegraphics[width=1\textwidth]{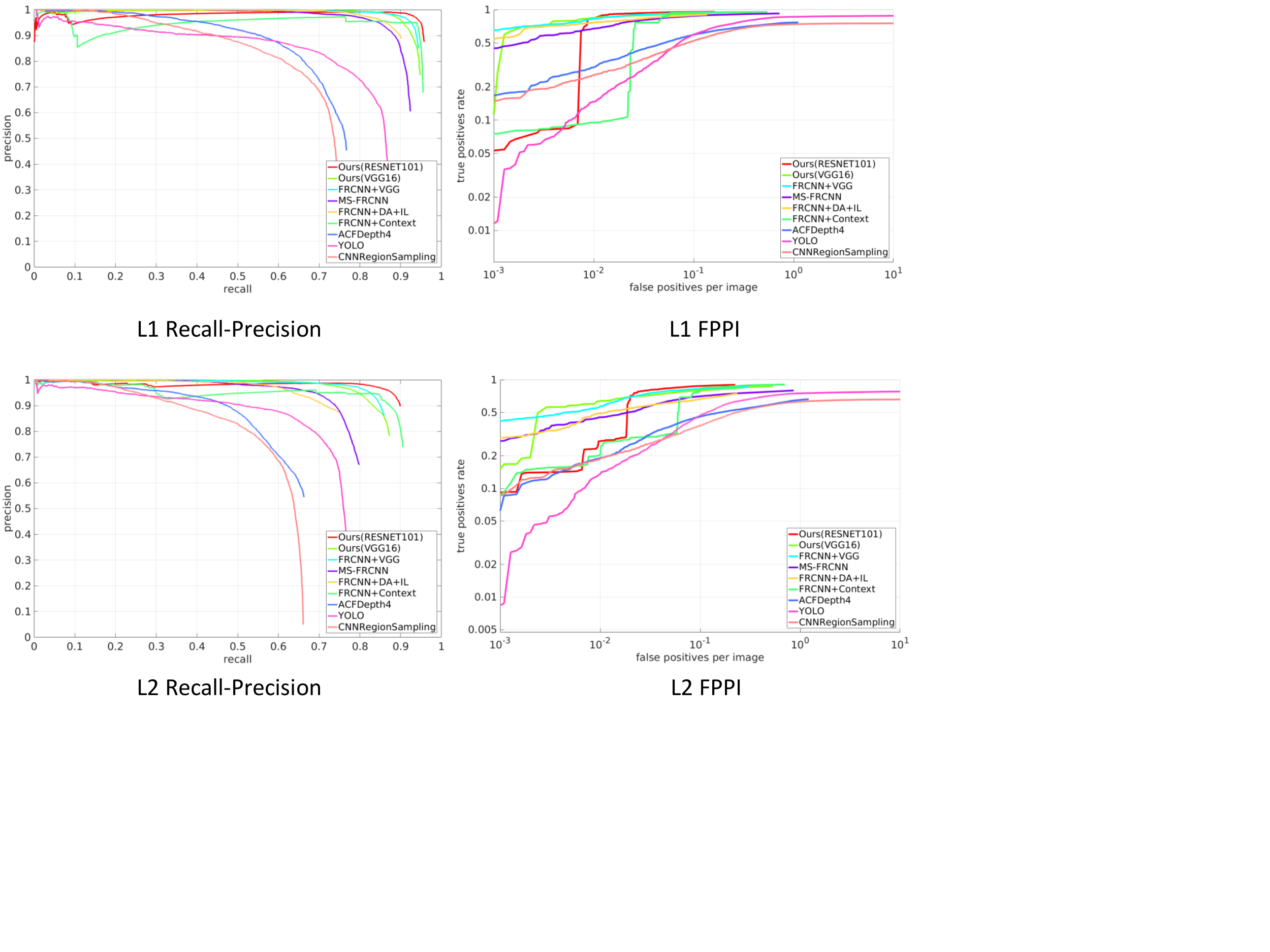}
\caption{ROC curves of hand detection on AP and AR measures obtained by FRCNN+VGG [22], MS-FRCNN [3], FRCNN+Context [22], FRCNN [17], ACF-Depth4 [18], YOLO [23], CNN-Region-Sampling [1] and our proposed method on VIVA dataset. (a) L1-AP, (b) L2-AP, (c) L1-AR, (d) L2-AR. Our method achieves the state-of-the-art hand detection results on this dataset.}
\label{fig:4}       
\end{center}
\end{figure*}


\begin{table}
\begin{center}
\caption{Performance of hand detection on AP, AR, FPS at both levels (L1 and L2) by FRCNN+VGG [22], MS-FRCNN [3], FRCNN [17], ACF-Depth4 [18], YOLO[23], CNNRegionSampling [1], MS-FRCNN [4] and our proposed method on VIVA dataset.}
\label{tab:1}       
\begin{tabular}{|l|l|l|l|}
\hline\noalign{\smallskip}
 & L1 AP/AR & L2 AP/AR & FPS \\
\hline
Ours (ResNets-101) & {\bfseries 95.7/95.3} & 91.3/{\bfseries 87.6} & 5.0 \\
\hline
Ours (VGG16) & 93.8/91.2 & 85.9/77.9 & 4.0 \\
\hline
MS-RFCN [4] & 95.1/86 & {\bfseries 94.5}/83.4 & 4.65 \\
\hline
FRCNN+VGG [22] & 93.9/91.5 & 85.2/77.8 & 6.33 \\
\hline
MS-FRCNN [3] & 90.8/84.13 & 77.6/65.1 & 4.65 \\
\hline
FRCNN+ VGG [22] & 89.5/86 & 73.4/64.4 & 4.9 \\
\hline
FRCNN [17] & 90.7/55.9 & 86.5/53.3 &  \\
\hline
ACF-Depth4 [18] & 70.1/53.8  & 60.1/40.4 &  \\
\hline
YOLO [23] & 76.4/46 & 69.5/39.1 & 3.5 \\
\hline
CNNRegionSampling [1] & 66.8/48.1 & 57.8/36.6 & 0.783 \\
\noalign{\smallskip}\hline
\end{tabular}
\end{center}
\end{table}


\begin{table}
\begin{center}
\caption{Performance of hand classification (left/right(L/R) and driver/passenger (D/P) on AP, AR and FPS obtained by ACF-Depth4[18], CNNRegionSampling [1], FRCNN+VGG [22], MS-RFCN [4] and our proposed method on VIVA dataset.}
\label{tab:1}       
\begin{tabular}{|l|l|l|l|}
\hline\noalign{\smallskip}
 & Left-Right Hand & Driver-Passenger Hand & FPS \\
\hline
Ours (ResNets-101) & {\bfseries 81.3/72.9} & {\bfseries 81.7/73.4} & 4.81 \\
\hline
Ours (VGG16) & 79.3/69.7 & 76.3/68.5 & 4.0 \\
\hline
MS-RFCN [4] & 75.3/69.8 & 70.9/65.6 & 4.65 \\
\hline
FRCNN+VGG [22] & 68.6/63.0 & 57.7/53.3 & 4.9 \\
\hline
ACF-Depth4 [18] & 47.5/33.7  & 43.1/30.5 &  11.6 \\
\hline
CNNRegionSampling [1] & 52.7/42.3 & 57.3/47.3 & 0.783 \\
\noalign{\smallskip}\hline
\end{tabular}
\end{center}
\end{table}

\begin{figure*}
\begin{center}
  \includegraphics[width=1\textwidth]{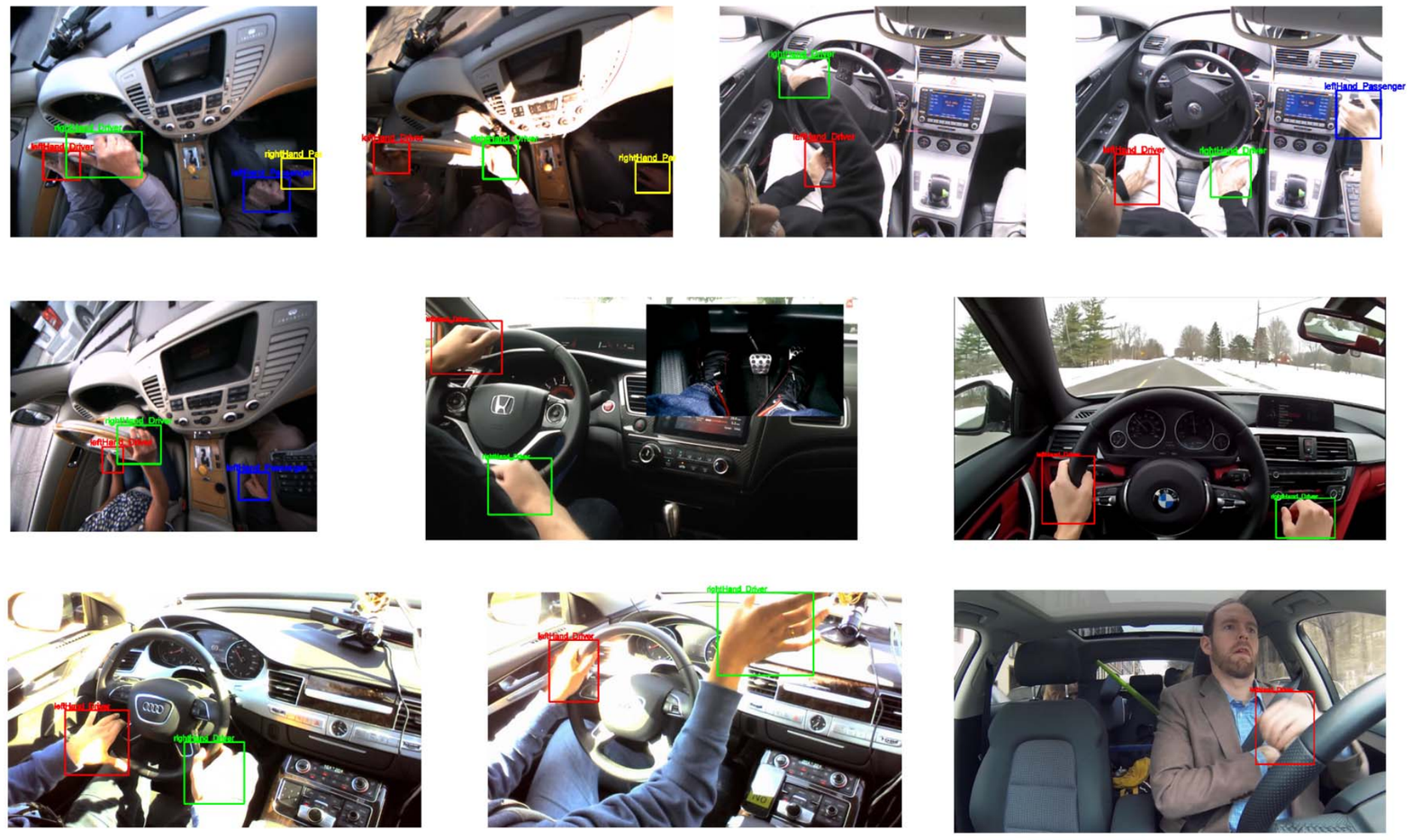}
\caption{Some examples of hand detection and classification results using our proposed method on VIVA dataset. Different colors represent different categories.}
\label{fig:4}       
\end{center}
\end{figure*}
\subsection{Experiments on Oxford Dataset}
Table 3 summaries the performance of our proposed method and the most recent methods on the Oxford dataset. Different approaches for hand detection on Oxford dataset have been proposed in [15]. From Figure 10, we can see that the proposed MSP-RFCN outperforms other methods with higher AP and AR.

Figure 8 demonstrates some examples of hand detection by our proposed MSP-RFCN method on Oxford dataset.


\begin{table}
\begin{center}
\caption{Performance of hand detection on AP and AR obtained by [15], MS-RFCN [4] and our proposed method on the Oxford dataset.}
\label{tab:1}       
\begin{tabular}{|l|l|l|}
\hline\noalign{\smallskip}
 & AP & AR \\
\hline
Mittal et al. [15] & 48.2 & {\bfseries 85.3} \\
\hline
MS-RFCN [4] & 75.1 & 77.8 \\
\hline
Ours (VGG16) & 76.2 & 79.6\\
\hline
Ours (ResNets-101) & {\bfseries 83.1} & 85.2 \\
\noalign{\smallskip}\hline
\end{tabular}
\end{center}
\end{table}

\begin{figure*}
\begin{center}
  \includegraphics[width=1\textwidth]{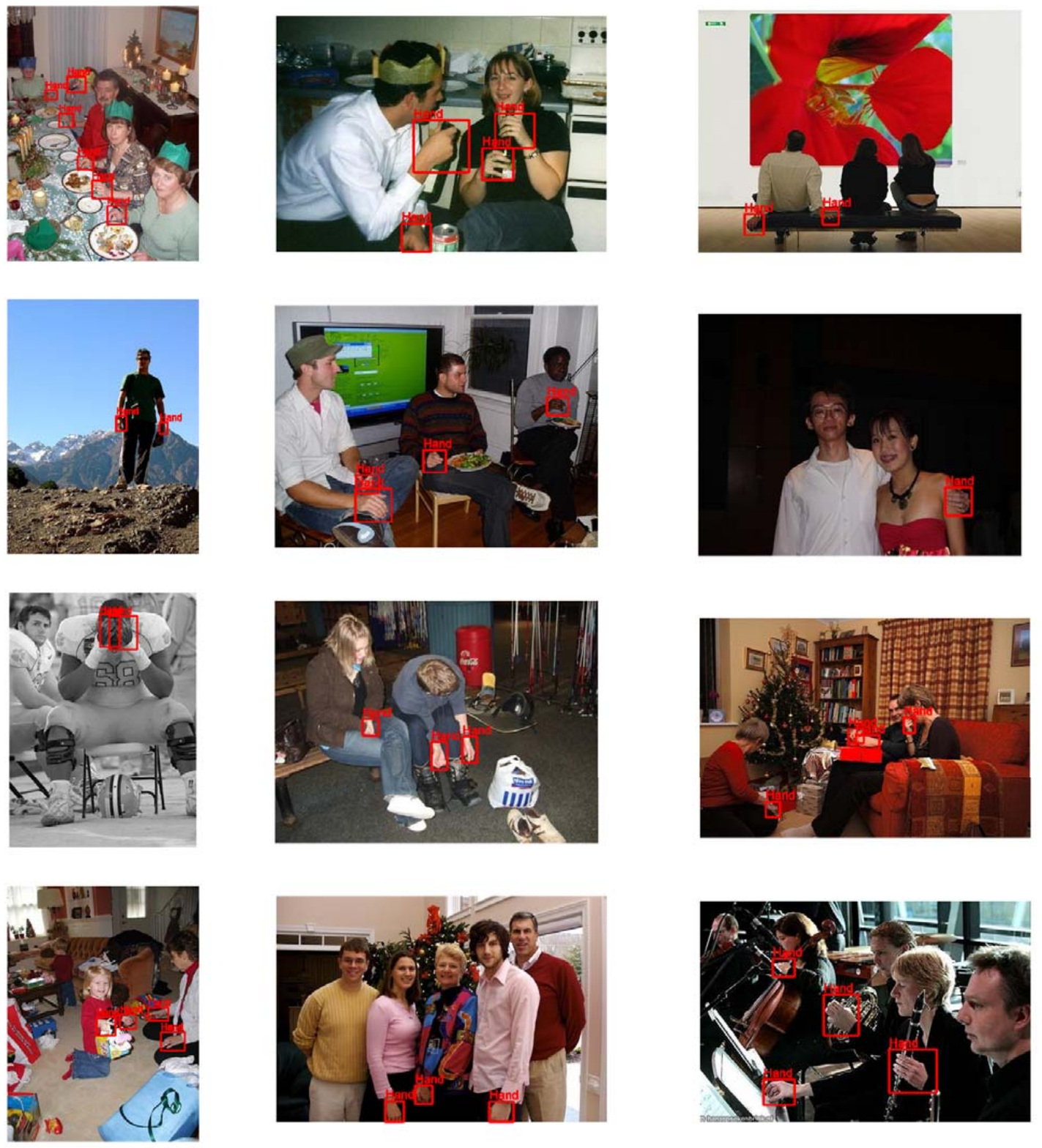}
\caption{Some examples of hand detection results using our proposed method on Oxford dataset.}
\label{fig:4}       
\end{center}
\end{figure*}
\subsection{Experiments on ARJ Dataset}
\begin{figure*}
\begin{center}
 \includegraphics[height=0.3\textheight]{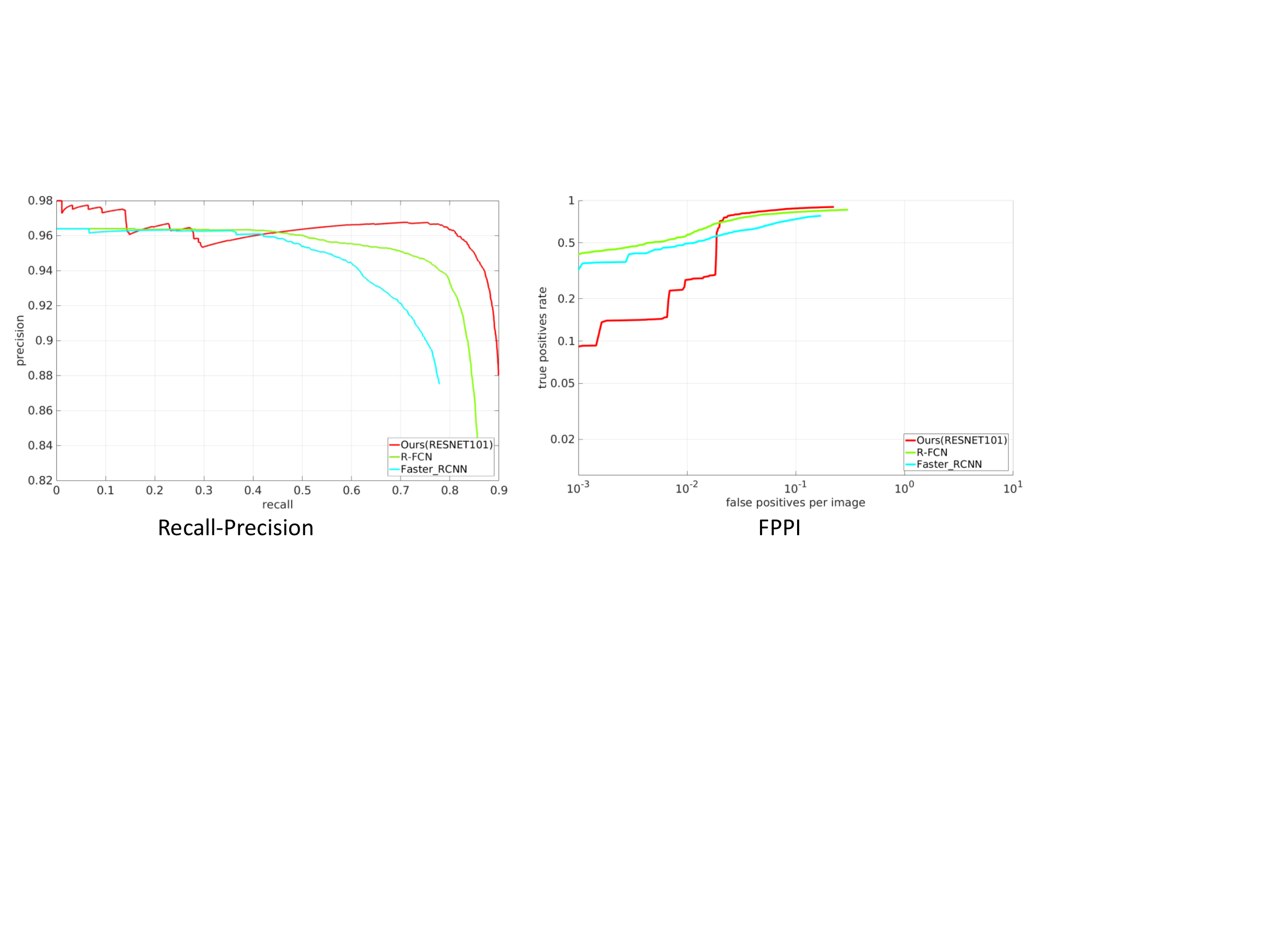}
\caption{ROC curves of hand detection on AP and AR on ARJ dataset.}
\label{fig:12}       
\end{center}
\end{figure*}

Figure 9 demonstrates the ROC and FPPI of our method and the most recent methods on ARJ dataset.

\begin{table}
\begin{center}
\caption{Performance of hand detection on AP and AR obtained by FRCNN method [10], R-FCN method [11], and our proposed method on ARJ dataset.}
\label{tab:1}       
\begin{tabular}{|l|l|l|}
\hline\noalign{\smallskip}
 & AP & AR \\
\hline
FRCNN [10] & 68.2 & 70.3 \\
\hline
R-FCN [11] & 73.1 & 75.4 \\
\hline
Ours (VGG16) & 81.2 & 85.7\\
\hline
Ours (ResNets-101) & {\bfseries 84.0} & {\bfseries 86.4} \\
\noalign{\smallskip}\hline
\end{tabular}
\end{center}
\end{table}

Table 4 summaries the performance of our proposed method and the most recent methods on ARJ dataset. Compare to FRCNN [10], R-FCN [11], our result achieves the state-of-the-art with 84.0\% of AP and 86.4\% of AR.

Figure 10 demonstrates some examples of hand detection by our proposed MSP-RFCN method on ARJ dataset.
%
\begin{figure*}
\begin{center}
  \includegraphics[width=1\textheight]{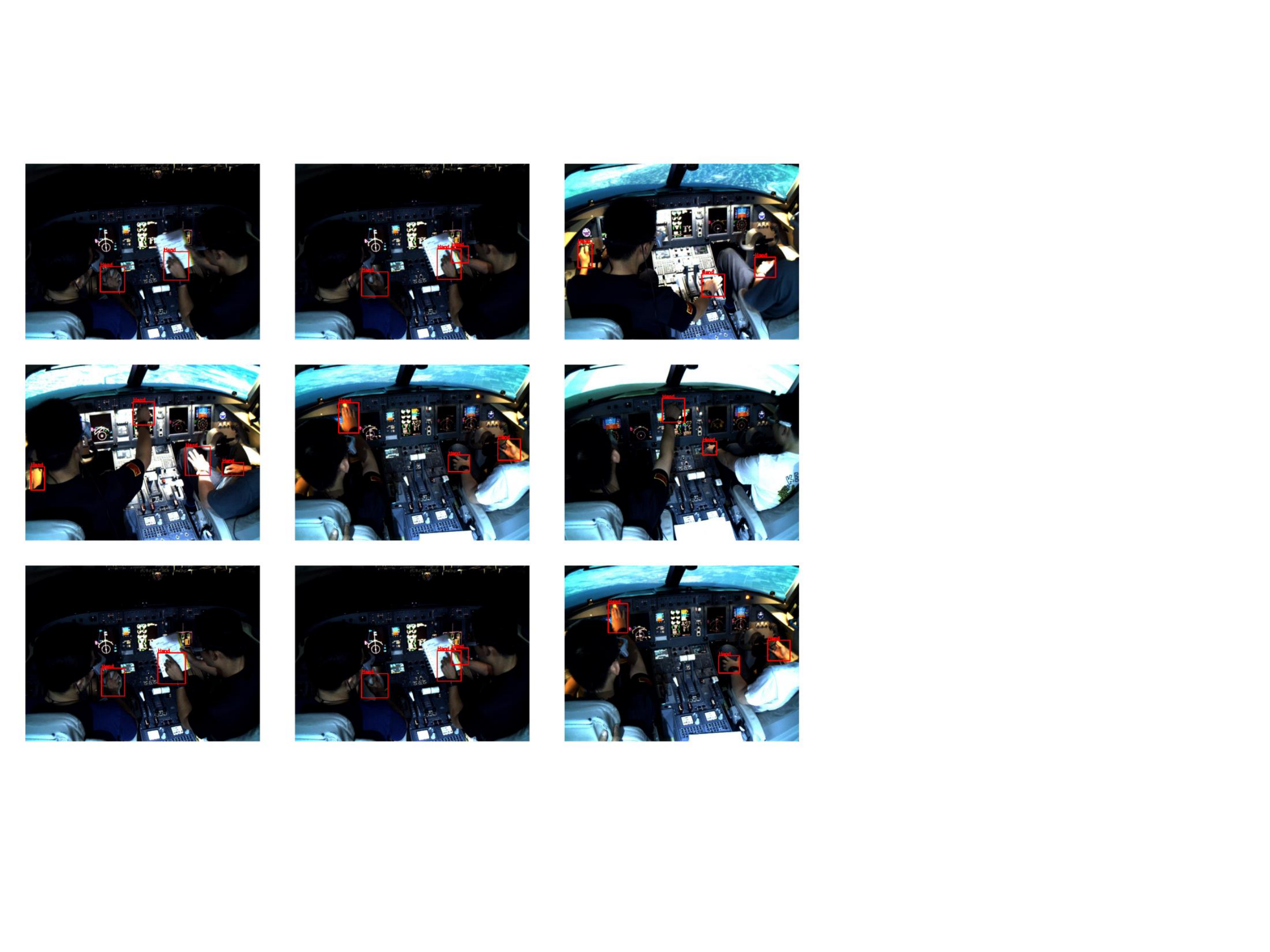}
\caption{Some examples of hand detection results using our proposed method on ARJ dataset.}
\label{fig:8}       
\end{center}
\end{figure*}
\subsection{Discussions}
Although our method has achieved favorable results in the three datasets, it is still far from perfect. For example, in Figure 9 (Precision-Recall), there exist sudden precision drops in our method. To improve our detector, we think there are two possible solutions. First, there are quantization errors in position-sensitive ROI pooling. In a recent work, ROI align [26] has been proposed as an effective solution to alleviate the adverse impact of quantization. Second, our multi-scale predictions have impact on small hand detection; large hand detection is not affected by multi-scale fusion. It could then be better to follow the approach in [27] to handle proposals with different scales in different scale fusion scheme.
\section{Conclusion}
In this paper, we present our proposed MSP-RFCN method to effectively and robustly detect and classify human hands from images collected in the wild under various challenging situations, e.g., heavy occlusions, illumination variations, low resolutions, etc. Our approach is based on the RFCN architecture to which we have added multi-scale prediction layers.
The method is evaluated on three challenging hand datasets, VIVA Challenge, Oxford hand dataset and ARJ dataset and compared against recent other detection methods, e.g., MS-RFCN, MS-FRCNN, etc. The experimental results demonstrate that our proposed method consistently achieves the state-of-the-art results on both hand detection and hand classification.

\ifCLASSOPTIONcaptionsoff
  \newpage
\fi

%

%
%
%




\end{document}